\documentclass[10pt,twocolumn,letterpaper]{article}
\usepackage[pagenumbers]{iccv} 

\newcommand{\gh}[1]{{\color{black} {#1}}}

\usepackage[utf8]{inputenc} 
\usepackage[T1]{fontenc}    
\usepackage{url}            
\usepackage{booktabs}       
\usepackage{amsfonts}       
\usepackage{nicefrac}       
\usepackage{microtype}      

\usepackage{graphicx}
\usepackage{amsmath}
\usepackage{amssymb}

\usepackage{diagbox}
\usepackage{multicol}
\usepackage{enumerate}
\usepackage{times}
\usepackage{epsfig}
\usepackage{threeparttable}
\usepackage{enumitem}
\usepackage{multirow}
\usepackage{color}
\usepackage{array}
\usepackage{setspace}
\usepackage{makecell}
 \usepackage{indentfirst}

\usepackage[colorlinks, linkcolor=red, citecolor=blue]{hyperref}

\title{LLaFEA: Frame-Event Complementary Fusion \\ for Fine-Grained Spatiotemporal Understanding in LMMs}

\author{Hanyu Zhou\textsuperscript{\rm 1}, Gim Hee Lee\textsuperscript{\rm 1}\\
  \textsuperscript{\rm 1} School of Computing, National University of Singapore\\
  {\tt\small {\{hy.zhou, gimhee.lee\}}@nus.edu.sg}
 }

\usepackage[sort&compress]{natbib}
\begin{document}
\maketitle

\begin{abstract}
Large multimodal models (LMMs) excel in scene understanding but struggle with fine-grained spatiotemporal reasoning due to weak alignment between linguistic and visual representations. Existing methods map textual positions and durations into the visual space encoded from frame-based videos, but suffer from temporal sparsity that limits language-vision temporal coordination. To address this issue, we introduce LLaFEA (Large Language and Frame-Event Assistant) to leverage event cameras for temporally dense perception and frame-event fusion. Our approach employs a cross-attention mechanism to integrate complementary spatial and temporal features, followed by self-attention matching for global spatio-temporal associations. We further embed textual position and duration tokens into the fused visual space to enhance fine-grained alignment. This unified framework ensures robust spatio-temporal coordinate alignment, enabling LMMs to interpret scenes at any position and any time. In addition, we construct a dataset of real-world frames-events with coordinate instructions and conduct extensive experiments to validate the effectiveness of the proposed method.

\end{abstract}

\section{Introduction}
\label{sec:intro}
Large multimodal models (LMMs)~\cite{alayrac2022flamingo,li2022blip,li2023blip2,liu2023visual} are designed to establish correspondence between language and other modalities (\eg, vision~\cite{radford2021learning} and audio~\cite{gong2022contrastive}) and have been widely applied to scene understanding tasks such as detailed captioning~\cite{wang2022git}, visual QA~\cite{li2022blip,li2023blip2}, scene grounding~\cite{kamath2021mdetr}, \etc. Recent advancements in language-vision LMMs (\eg, LLaVA~\cite{liu2023llava} and PaLI~\cite{chen2023pali}) have improved spatial and temporal scene understanding but still face challenges in fine-grained spatiotemporal reasoning. This difficulty arises because fine-grained understanding requires precise spatiotemporal coordinate alignment (\ie, position and duration) between linguistic and visual representations, yet the vast range of possible position-duration combinations complicates feature correspondence modeling. In this work, our objective is to improve the spatio-temporal alignment between linguistic and visual representations, thus strengthening the capability of LLMs for fine-grained spatiotemporal understanding.


\begin{figure}
  \setlength{\abovecaptionskip}{5pt}
  \setlength{\belowcaptionskip}{-5pt}
  \centering
   \includegraphics[width=0.99\linewidth]{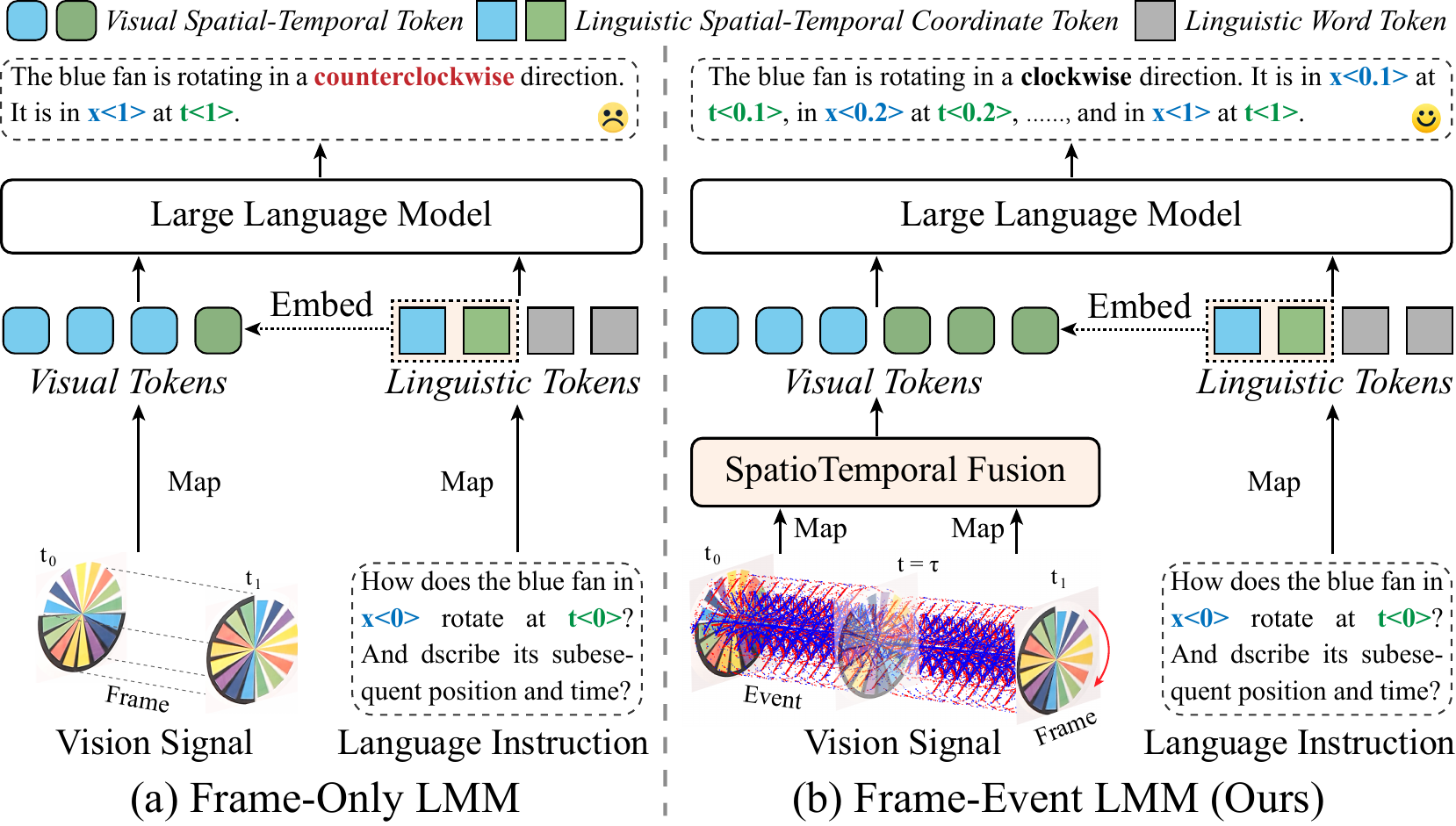}
   \caption{
   Illustration of two LMM paradigms for fine-grained spatiotemporal understanding. This task relies on aligning linguistic and visual representations in spatiotemporal coordinates. Frame-only LMMs embed these coordinates into visual tokens from frame videos, but suffer from temporal sparsity that weakens alignment. We incorporate a temporally dense event camera to enhance spatiotemporal representation for LMMs to interpret scenes more accurately at any position and time.
   } 
   \label{Fig:Paradigm}
\end{figure}

Most existing language-vision LMM methods perform point-to-point mapping of textual position and duration onto the visual space from frame-based videos in Fig. \ref{Fig:Paradigm} (a). For example, MDETR~\cite{kamath2021mdetr} and GIT~\cite{wang2022git} align textual descriptions with specific frames to ground objects in video scenes, while BLIP-2~\cite{li2023blip2} and LLaVA~\cite{liu2023llava} extend vision-language pretraining to incorporate temporal information by leveraging sequential frames. Although these methods have shown promising results in fine-grained spatiotemporal understanding, their performance remains constrained by the temporal sparsity of frame-based video capture. First, the low frame rate of traditional cameras introduces discontinuous motion, making it difficult for LMMs to infer the fine-grained temporal dynamics of a scene based on textual duration. Second, textual timestamps may not correspond precisely to existing frames, leading to temporal misalignment between language and vision. These limitations weaken the spatiotemporal coordination of linguistic and visual representations.
Improving the spatiotemporal density of visual representations is thus crucial for fine-grained spatiotemporal understanding.


\gh{To address this issue, we introduce event cameras with high temporal resolution to mitigate the temporal limitations of frame cameras and leverage their complementary nature to strengthen LMMs with spatiotemporal dense understanding in Fig. \ref{Fig:Paradigm} (b). 
From an imaging perspective, frame cameras capture the global appearance of an entire scene through uniform exposure over a fixed time interval, while event cameras asynchronously trigger event signals at local boundaries in response to brightness changes. This highlights a clear complementary relationship: frame-based imaging provides spatially dense but temporally sparse appearance information, and event-based imaging offers spatially sparse but temporally dense boundary details. From a feature perspective, we observe that despite their distinct capture mechanisms, frame and event modalities share a homogeneous spatiotemporal feature distribution for the same scene content. This suggests that their combined representation can effectively model the entire scene while preserving both spatial and temporal information.  Motivated by these insights, we propose a hierarchical fusion framework to integrate complementary spatio-temporal information from frame and event modalities to boost spatio-temporal density of visual representations.
}

\begin{figure*}
  \setlength{\abovecaptionskip}{5pt}
  \setlength{\belowcaptionskip}{-5pt}
  \centering
   \includegraphics[width=0.98\linewidth]{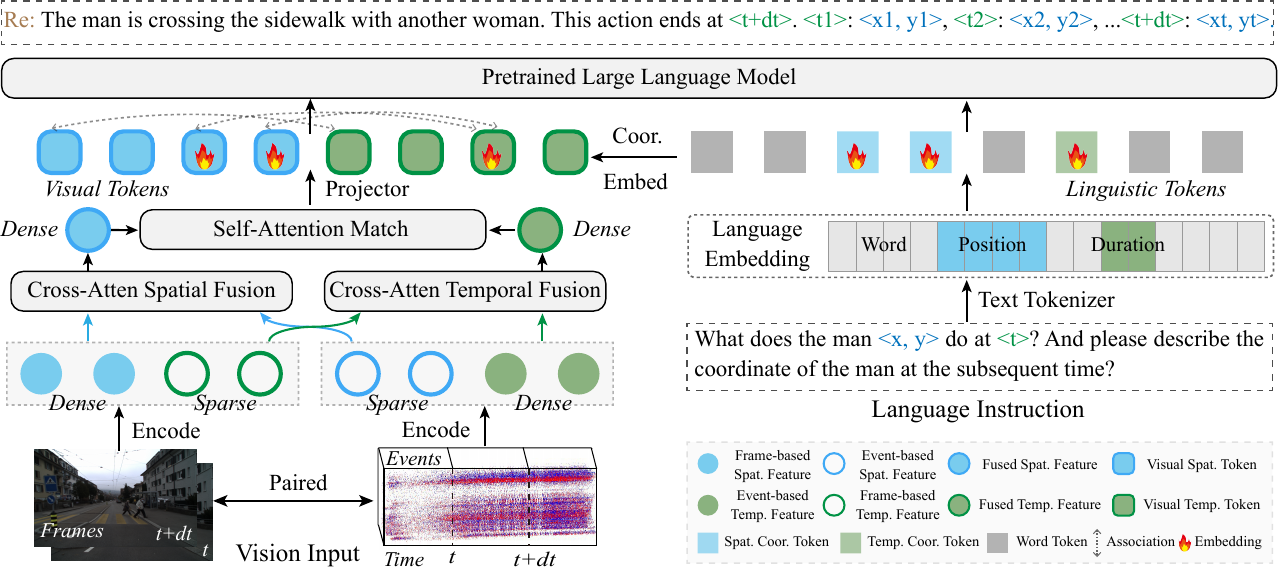}
   \caption{
   \gh{Our \emph{LLaFEA} comprises visual frame-event spatiotemporal fusion and language-vision coordinate alignment. The first stage fuses spatial-dense, temporal-sparse frame features with spatial-sparse, temporal-dense event features to generate spatiotemporal-dense visual tokens. The second stage embeds spatiotemporal coordinate tokens from the language embedding into the fused visual tokens, which are then processed by the LLM for fine-grained understanding.}
   } 
   \label{Fig:Framework}
\end{figure*}

\gh{In this work, we propose LLaFEA (Large Language and Frame-Event Assistant) in Fig. \ref{Fig:Framework}, a novel framework designed for fine-grained spatiotemporal-dense understanding by integrating visual frame-event spatiotemporal fusion and language-vision coordinate alignment. This approach strengthens the ability of LLMs to interpret scene content with precise spatial and temporal awareness. To achieve effective visual frame-event spatiotemporal fusion, we first encode frame videos and event streams into their respective spatial and temporal feature representations. We then introduce a hierarchical fusion-matching process to integrate the complementary strengths of both modalities. In the fusion step, we apply cross-attention spatial fusion and cross-attention temporal fusion, both utilizing a shared transformer architecture. The spatial fusion process treats frame-based spatially dense features as the primary representation and integrates event-based spatially sparse features. Conversely, the temporal fusion process prioritizes event-based temporally dense features to enrich frame-based temporally sparse features. In the matching step, we implement a self-attention matching strategy to establish global associations between the fused spatial and temporal features, ensuring a more comprehensive spatiotemporal representation.

For language-vision coordinate alignment, we first introduce an MLP-based projection layer to transform the fused spatiotemporal visual features into visual tokens within the language space. These tokens are then enriched with textual spatiotemporal coordinate tokens before being processed by the LLM. This process facilitates precise fine-grained spatiotemporal reasoning by aligning linguistic and visual representations at both spatial and temporal levels.
By unifying hierarchical frame-event fusion with language-vision alignment, our LLaFEA improves spatiotemporal-dense coordinate alignment, facilitating LMMs to understand and reason about scene content at any position and time. Additionally, we construct a real-world frame-event dataset with spatiotemporal coordinate annotations to enable fine-tuning and validation of our approach for better fine-grained spatiotemporal understanding.
}

\gh{Our main contributions are summarized as follows:}
\begin{itemize}[leftmargin=10pt]

\item \gh{We propose a novel large language and frame-event model that integrates event cameras to complement frame cameras in visual representation, allowing LMMs to achieve fine-grained spatiotemporally dense understanding.}

\item \gh{We identify the complementary and structurally consistent properties of spatiotemporal features between frame and event modalities. This insight drives the design of a hierarchical fusion framework to improve spatiotemporal-dense coordinate alignment between linguistic and visual representations.}


\item \gh{We construct a real-world frame-event dataset with spatiotemporal coordinate annotations and conduct extensive experiments to demonstrate the effectiveness of our approach under diverse conditions, including high-dynamic and low-light environments.}
\end{itemize}

\section{Related Work}
\label{sec:related_work}
\noindent
\textbf{Multimodal Scene Understanding.}
\gh{Multimodal scene understanding aims to establish the representation correspondence among various modalities and integrate their knowledge to capture the spatiotemporal characteristics of a scene. Traditional multimodal understanding methods rely on large amounts of annotated multimodal data to train specialized deep networks for different spatiotemporal knowledge domains, such as semantic segmentation~\cite{chen2017deeplab}, depth estimation~\cite{eigen2014depth}, and motion analysis~\cite{dosovitskiy2015flownet}. However, these methods are limited in generalization and reasoning capabilities as they primarily learn statistical patterns from training data. Consequently, they struggle to adapt to new tasks and infer logical relationships within a scene. In contrast, the versatility and reasoning power of LLMs~\cite{brown2020gpt3} have inspired the development of LMMs such as LLaVA~\cite{liu2023visual} and Flamingo~\cite{alayrac2022flamingo}. These LMMs align images and text within a unified representation space to understand and reason about spatiotemporal knowledge, enabling the model to perform diverse tasks with simple prompting or fine-tuning such as image captioning~\cite{wang2022git} and visual question answering (VQA)~\cite{li2022blip,li2023blip2}. For example, LLaVA-1.5~\cite{liu2023improved} enhances image-text alignment by integrating richer vision-language representations for more precise multimodal scene understanding.
}

\noindent
\textbf{LMMs with Fine-Grained Understanding.}
Language-vision large multimodal models (LMMs) aim to bridge the gap between textual descriptions and visual representations for spatiotemporal scene understanding. The key challenge in achieving fine-grained spatiotemporal understanding is aligning the spatiotemporal coordinates of linguistic and visual representations. An intuitive approach is to perform a point-to-point mapping of textual spatial positions and temporal durations onto the visual space encoded from video frames. For example, temporal grounding models such as Moment-DETR~\cite{lei2021momentdetr}, and VideoCLIP~\cite{xu2021videoclip} employ transformer-based architectures to directly link textual event descriptions with corresponding temporal segments in videos. Additionally, CLIP-based video models, such as CLIP4Clip~\cite{luo2022clip4clip} and VATT~\cite{akbari2021vatt}, use contrastive learning to refine video-language alignment at a fine-grained level.

Although these methods have made notable progress in fine-grained spatiotemporal understanding, their performance remains constrained by the temporal sparsity introduced by the low frame rate of frame-based cameras. There are two reasons to this limitation: 1) Videos captured with frame-based cameras often exhibit temporally discontinuous motion, especially in high-dynamic scenes, making it difficult for LMMs to infer the fine-grained variations of dynamic patterns across the temporal dimension. This discontinuity complicates the ability of the model to establish a coherent correspondence between visual motion patterns and textual durations. 2) Textual timestamps may not be precisely aligned with existing video frames, causing temporal token misalignment and inconsistencies in video-language mapping.
We focus on improving the spatiotemporal density of visual representation, thus ensuring the language-vision coordinate alignment of LMMs.


\noindent
\textbf{Event-Based Multimodal Model.}
Compared to frame imaging with uniform exposure, an event camera is a neuromorphic vision sensor with high temporal resolution and high dynamic range, making it a promising tool for enhancing frame-based cameras in spatiotemporal densification of visual representations~\cite{gallego2020event, posch2014retinomorphic}. An approach to leveraging event cameras is to perform video interpolation~\cite{tulyakov2018softmax, shedligeri2019photorealistic, scheerlinck2020fast, rebecq2019e2vid} using event data, generating high-frame-rate videos that can then be processed by LMMs. However, although this strategy improves visualization density, it does not necessarily improve spatiotemporal feature alignment between linguistic and visual representations. Additionally, interpolation artifacts and hallucinated frames in generated videos can introduce inconsistencies that negatively impact model predictions. A more effective solution is to model the feature interaction between frame-based and event-based visual representations for spatiotemporal understanding in LMMs. For example, event-frame fusion networks such as EvT~\cite{gu2022evt}, E2VID~\cite{rebecq2019e2vid}, and EST~\cite{zhang2023est} integrate event and frame features to boost motion representation and low-light robustness. Additionally, transformer-based fusion frameworks such as EFormer~\cite{wang2023eformer} and EST-Former~\cite{xu2023estformer} refine event-frame feature alignment, leading to temporally dense and semantically rich representations. Although these methods have advanced event-camera-assisted spatiotemporal understanding in LMMs, they do not fully use the complementary advantages of frame and event modalities for fine-grained understanding. We pioneer the exploration of complementary spatial-temporal information from frame and event modalities to enhance visual representation density, enabling LMMs to interpret scene content at any position and any time.

\section{Large Language and Frame-Event Assistant}
\label{sec:method}

\noindent \textbf{Overview.} 
Fig.~\ref{Fig:Framework} shows our framework with two key stages: 1) \textbf{\textit{Visual Frame-Event Spatiotemporal Fusion}}, where complementary spatial and temporal features from both modalities are extracted, transformed, and fused with cross-attention and self-attention mechanisms. 2) \textbf{\textit{Language-Vision Coordinate Alignment}}, where fused features are projected to the language embedding space and aligned with textual spatiotemporal coordinates. This integration enables LMMs to achieve more precise scene understanding at any position and time, significantly improving spatiotemporal coherence in vision-language tasks.

\vspace{2mm}
\noindent \textbf{Our Framework.}
Given a video sequence \( I_v \) and its corresponding event stream \( I_e \), we employ a frame encoder \( F_v \) and an event encoder \( F_e \) to extract visual features \( f_v \) and \( f_e \).

\vspace{1mm}
\noindent \textbf{1) \textit{Visual Frame-Event Spatiotemporal Fusion.}} 
We utilize a spatiotemporal mapper to transform the extracted visual features into spatiotemporal representations: [$f_{vs}$, $f_{vt}$] and [$f_{es}$, $f_{et}$]. To further integrate cross-modal spatial and temporal information, we employ a cross-attention fusion mechanism:
\begin{equation}
  \setlength\abovedisplayskip{3pt}
  \setlength\belowdisplayskip{3pt}
\begin{aligned}
 f_s = \operatorname{CAttn}(f_{vs}, f_{es}), \quad f_t = \operatorname{CAttn}(f_{vt}, f_{et}),
  \label{eq:cross_attention}
\end{aligned}
\end{equation}
where $f_s$ and $f_t$ are spatial-dense features and temporal-dense features, respectively. Subsequently, we propose a self-attention matching strategy to learn the global association of the fused features: $f_{st} = \operatorname{SAttn}(f_{s}, f_{t}).$

\vspace{1mm}
\noindent \textbf{2) \textit{Language-Vision Coordinate Alignment.}}
The features from the previous stage are projected into language embedding space using a multi-layer perception: $v_{st}=\operatorname{MLP}(f_{st})$,
where $v_{st}$ denotes visual tokens. We fuse textual position and duration into the spatiotemporal coordinate tokens $\tilde{v}$, which are embedded into the visual tokens: $\tilde{v_{st}} = v_{st} + \tilde{v}$.

\vspace{1mm}
\noindent \textbf{Remarks.} The output \( \tilde{v_{st}} \) is a spatiotemporal dense visual representation with language-aligned coordinates. Subsequently, the LLM~\cite{liu2023improved} leverages these visual, coordinate, and word tokens to improve spatiotemporal understanding. Our unified framework integrates frame-event fusion and language-vision alignment to enable fine-grained spatiotemporal reasoning in LMMs. The following sections detail the motivations and design of each stage.



\subsection{Visual Frame-Event Spatiotemporal Fusion}
The density of visual representation is crucial for fine-grained spatiotemporal understanding in LMMs. However, due to the inherent limitations of frame cameras, visual representations exhibit temporal sparsity that causes misalignment between linguistic and visual temporal coordinates to impede fine-grained understanding. We thus incorporate event cameras with high temporal resolution to complement frame cameras and enrich visual representation. However, this introduces two challenges: what and how to fuse.

\begin{figure}
  \setlength{\abovecaptionskip}{5pt}
  \setlength{\belowcaptionskip}{-7pt}
  \centering
   \includegraphics[width=0.99\linewidth]{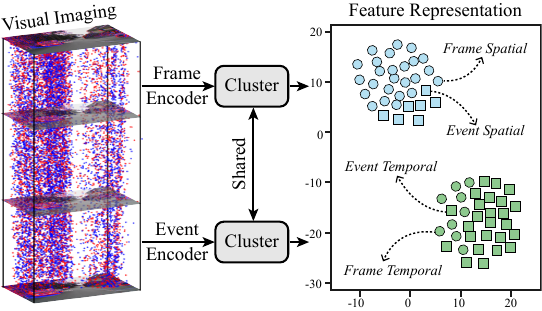}
   \caption{
   \gh{Visual imaging and feature representation of frames and events. In visual imaging, frames capture a spatially dense and temporally sparse global appearance. Events detect spatially sparse and temporally dense local boundaries. In feature representation, spatial and temporal feature distributions exhibit strong similarities between the two modalities.}
   } 
   \label{Fig:Analysis}
\end{figure}

\vspace{1mm}
\noindent
\textbf{Dense \emph{v.s} Sparse Spatiotemporal Knowledge.}
To leverage the complementary knowledge of frame and event modalities, we analyze their differences from two perspectives: imaging mechanism and feature representation.
Frame cameras capture global appearance with spatial density but suffer from temporal sparsity as exposure time depends on illumination. In contrast, event cameras operate asynchronously, detecting local motion boundaries with temporal density but spatial sparsity. As shown in Fig.~\ref{Fig:Analysis}, this complementarity enhances spatiotemporal fusion in visual representation. However, direct fusion is challenging since the domain gap between frame and event imaging. We thus explore a latent space with reduced domain disparity. To this end, we introduce a spatiotemporal mapper using a patch-level inner product operator, transforming visual features \( f_v \) and \( f_e \) into spatial \([f_{vs}, f_{vt}]\) and temporal \([f_{es}, f_{et}]\) representations that preserves their correlations.  
To assess the feature consistency, we cluster and visualize spatiotemporal features of dynamic targets using t-SNE~\cite{vanDerMaaten2008tsne} in Fig.~\ref{Fig:Analysis}, revealing minimal domain gap between frame and event modalities. This motivates a joint fusion strategy in spatial and temporal feature spaces to boost spatiotemporal density for more precise scene understanding in LMMs.

\vspace{1mm}
\noindent
\textbf{Cross-Attention Fusion.}
The core of complementary knowledge fusion lies in modeling the dependency between frame and event modalities, motivating us to adopt an attention-based cross-modal fusion mechanism. Since spatial and temporal features must be estimated separately, we design a cross-attention spatial fusion and a cross-attention temporal fusion that utilize similar transformer-based architectures. For cross-attention spatial fusion, we use frame-based spatial-dense features as the primary representation and fuse them with event-based spatial-sparse features as:
\begin{equation}\small
  \setlength\abovedisplayskip{2pt}
  \setlength\belowdisplayskip{2pt}
\begin{aligned}
  &q = w_qf_{vs}, \quad k = w_kf_{es}, \quad v = w_vf_{es},\\
  &A = \operatorname{softmax}(qk^{t}/\sqrt{d}), \quad \hat{f_s} = Av, \\
  &f_s = \operatorname{LayerNorm}(f_{vs}+\hat{f_s}),
  \label{eq:transformer_cross}
\end{aligned}
\end{equation}
where \( d \) represents the feature dimension, and \( w \) is a trainable weight. The output \( f_s \) is the spatial-dense visual feature. For cross-attention temporal fusion, we apply the same fusion strategy as in Eq.~\ref{eq:transformer_cross}, using event-based temporal-dense features as the primary representation to integrate frame-based temporal-sparse features, resulting in a temporal-dense visual feature \( f_t \).  
By using both imaging mechanisms and feature representations, the fused spatial and temporal features preserve dense and continuous appearance information while enhancing boundary details to facilitate fine-grained spatiotemporal understanding.


\vspace{1mm}
\noindent
\textbf{Self-Attention Matching.}
Although cross-attention fusion generates dense spatial and temporal features from frame and event modalities, these features remain independent without direct interaction. This absence of association hinders LMMs from effectively reasoning about fine-grained spatiotemporal relationships in a scene. We thus introduce an attention-based spatiotemporal matching strategy. Specifically, we adopt a transformer architecture as the matching backbone to capture global associations between spatial and temporal features as:
\begin{equation}\small
  \setlength\abovedisplayskip{2pt}
  \setlength\belowdisplayskip{2pt}
\begin{aligned}
  &f = \operatorname{Concat}(f_s, f_t), \quad q=w_qf, \quad 
k=w_kf, \quad v=w_vf,\\
  &A=\operatorname{softmax}(qk^{T}/\sqrt{d}), \quad \hat{f}=Av,\\
  &\hat{f_s} = \hat{f}[:N], \quad \hat{f_t} = \hat{f}[N:], \quad f_{st} = \operatorname{Concat}(\hat{f_s}, \hat{f_t}),
  \label{eq:transformer_self}
\end{aligned}
\end{equation}
where \( N \) represents the number of spatial features. The output \( f_{st} \) is the final spatiotemporal-dense feature, serving as the foundation for aligning spatiotemporal coordinates between linguistic and visual representations.


\subsection{Language-Vision Coordinate Alignment}
Frame-event complementary fusion involves feature interaction within visual representation, while fine-grained spatiotemporal understanding requires aligning the spatiotemporal coordinates of linguistic and visual representations, including position and duration. To bridge this gap, we embed textual position and duration into the fused visual space, enabling the model to learn fine-grained spatiotemporal correspondences between the two modalities.

\vspace{1mm}
\noindent
\textbf{Spatiotemporal Coordinate Token.}
Since the input of large language model requires text-like tokens, we first build a multi-layer perception to project the fused visual features into language embedding tokens, namely visual tokens $v_{st}$. Next, we tokenize text position and duration into spatiotemporal coordinate tokens $p$, $t$, which are converted to the same space as the visual tokens as follows: $v_p = w_p \cdot p, v_t = w_t \cdot t$,
where $w_p$, $w_t$ are the learnable linear mapping matrix. We further unify the two tokens to the spatiotemporal token: $\tilde{v} = v_p + v_t$.
This allows the spatiotemporal token $\tilde{v}$ to encapsulate both spatial position and temporal duration, acting as an intermediate bridge between linguistic and visual representations for fine-grained spatiotemporal understanding.

\vspace{1mm}
\noindent
\textbf{Coordinate Embedding.}
To establish spatiotemporal correspondence between linguistic and visual representations, we embed the spatiotemporal coordinate token \( \tilde{v} \) into the visual token \( v_{st} \) using a learnable fusion strategy: $\tilde{v_{st}} = v_{st} + \alpha\tilde{v}$.
Here, \( \alpha \) is a learnable weight parameter, and \( \tilde{v_{st}} \) represents the coordinate-aligned visual token incorporating position and duration information. We apply $\tilde{v_{st}}$ to all visual tokens and concatenate the fused visual tokens with text tokens for the LLM to reason. Within the framework, frame-event fusion enhances the spatiotemporal density of visual representation, while language-vision alignment reinforces spatiotemporal coordinate alignment, jointly enabling the LMM to achieve fine-grained spatiotemporal dense understanding.


\begin{table}\footnotesize
    \setlength{\abovecaptionskip}{5pt}
    \setlength\tabcolsep{4pt}
    \setlength{\belowcaptionskip}{-7pt}
  \centering
  \renewcommand\arraystretch{1.1}
  \begin{tabular}{cccc}
    \Xhline{1pt}
      \multicolumn{1}{c}{\multirow{1}{*}{Training stage}} &
      \multicolumn{1}{c}{\multirow{1}{*}{Task}} & \multicolumn{1}{c}{\multirow{1}{*}{Sample}}&\multicolumn{1}{c}{\multirow{1}{*}{Dataset}} \\
      \hline 
      S1: Content Alignment & DC & 1.60 M & SynEvent-ST \\
      S2: Fusion and Matching & DC, VQA & 2.56 M & SynEvent-ST \\
      S3: Coordinate Alignment & STG & 640 K & SynEvent-ST \\
      S4: Instruction Fine-tuning & STG & 74 K & RealEvent-ST \\
       \Xhline{1pt}
  \end{tabular}
  \caption{Overview of training pipelines and datasets.} 
   \label{Tab:Dataset}
\end{table}

\section{Datasets and Training Pipelines}
\label{sec:training}

\subsection{Datasets}
\gh{Given the absence of event datasets for spatiotemporal understanding tasks, we create a synthetic event dataset to initialize multimodal understanding and a real event dataset for instruction fine-tuning in Table \ref{Tab:Dataset}.}

\vspace{1mm}
\noindent
\textbf{LLaFEA-SynEvent-ST.} 
\gh{LMMs require a large volume of vision-language sample pairs to develop multimodal spatiotemporal understanding. To this end, we integrate standard spatiotemporal understanding datasets~\cite{bain2021frozen, chen2024panda, wang2023internvid, wang2024internvideo2, luo2023valley, maaz2023video} and employ the v2e model~\cite{hu2021v2e} to simulate event data from frame images, generating a new multimodal dataset with $3.2$M samples. Among the dataset, $80\%$ focus on detailed captioning and visual question answering for spatiotemporal understanding, while $20\%$ further increases spatial-temporal grounding tasks for fine-grained spatiotemporal understanding.
}

\begin{table*}\footnotesize
    \setlength{\abovecaptionskip}{5pt}
    \setlength\tabcolsep{4pt}
    \setlength{\belowcaptionskip}{-5pt}
  \centering
  \renewcommand\arraystretch{1.02}
  \begin{tabular}{cccccccccccc}
    \Xhline{1pt}
      \multicolumn{2}{c|}{\multirow{2}{*}{Method}} &
      \multicolumn{1}{c|}{\multirow{1}{*}{Video-ChatGPT}} & \multicolumn{2}{c|}{\multirow{1}{*}{Video-LLaVA}} &
      \multicolumn{2}{c|}{\multirow{1}{*}{GroundingGPT}} &
      \multicolumn{2}{c|}{\multirow{1}{*}{Grounded-VideoLLM}} &
      \multicolumn{2}{c|}{\multirow{1}{*}{VTimeLLM}} & 
      \multicolumn{1}{c}{\multirow{2}{*}{Ours}}\\
      \cline{3-11}
      \multicolumn{2}{c|}{\multirow{1}{*}{}} &
      \multicolumn{1}{c|}{\multirow{1}{*}{--}} &
      \multicolumn{1}{c|}{\multirow{1}{*}{--}} &
      \multicolumn{1}{c|}{\multirow{1}{*}{w/ FI}} &
      \multicolumn{1}{c|}{\multirow{1}{*}{--}} &
      \multicolumn{1}{c|}{\multirow{1}{*}{w/ FI}} &
      \multicolumn{1}{c|}{\multirow{1}{*}{--}} &
      \multicolumn{1}{c|}{\multirow{1}{*}{w/ FI}} &
      \multicolumn{1}{c|}{\multirow{1}{*}{--}} &
      \multicolumn{1}{c|}{\multirow{1}{*}{w/ FI}} &
      \multicolumn{1}{c}{\multirow{1}{*}{}} \\
      \hline
      \multicolumn{2}{c|}{\multirow{1}{*}{Vision input}} &
      \multicolumn{1}{c|}{\multirow{1}{*}{Frame}} &
      \multicolumn{1}{c|}{\multirow{1}{*}{Frame}} &
      \multicolumn{1}{c|}{\multirow{1}{*}{Frame+Event}} &
      \multicolumn{1}{c|}{\multirow{1}{*}{Frame}} &
      \multicolumn{1}{c|}{\multirow{1}{*}{Frame+Event}} &
      \multicolumn{1}{c|}{\multirow{1}{*}{Frame}} &
      \multicolumn{1}{c|}{\multirow{1}{*}{Frame+Event}} &
      \multicolumn{1}{c|}{\multirow{1}{*}{Frame}} &
      \multicolumn{1}{c|}{\multirow{1}{*}{Frame+Event}} &
      \multicolumn{1}{c}{\multirow{1}{*}{Frame+Event}} \\
      \hline
      \multicolumn{1}{c|}{\multirow{4}{*}{\makecell{Syn\\Event\\-ST}}} &
      \multicolumn{1}{c|}{\multirow{1}{*}{DC}} &
      \multicolumn{1}{c|}{\multirow{1}{*}{2.51}} &
      \multicolumn{1}{c|}{\multirow{1}{*}{2.14}} &
      \multicolumn{1}{c|}{\multirow{1}{*}{2.75}} &
      \multicolumn{1}{c|}{\multirow{1}{*}{2.58}} &
      \multicolumn{1}{c|}{\multirow{1}{*}{2.91}} &
      \multicolumn{1}{c|}{\multirow{1}{*}{2.64}} &
      \multicolumn{1}{c|}{\multirow{1}{*}{3.27}} &
      \multicolumn{1}{c|}{\multirow{1}{*}{2.83}} &
      \multicolumn{1}{c|}{\multirow{1}{*}{3.30}} &
      \multicolumn{1}{c}{\multirow{1}{*}{\textbf{3.94}}} \\
      \multicolumn{1}{c|}{\multirow{1}{*}{}} &
      \multicolumn{1}{c|}{\multirow{1}{*}{VQA}} &
      \multicolumn{1}{c|}{\multirow{1}{*}{2.80}} &
      \multicolumn{1}{c|}{\multirow{1}{*}{2.56}} &
      \multicolumn{1}{c|}{\multirow{1}{*}{3.12}} &
      \multicolumn{1}{c|}{\multirow{1}{*}{2.85}} &
      \multicolumn{1}{c|}{\multirow{1}{*}{3.24}} &
      \multicolumn{1}{c|}{\multirow{1}{*}{2.96}} &
      \multicolumn{1}{c|}{\multirow{1}{*}{3.63}} &
      \multicolumn{1}{c|}{\multirow{1}{*}{3.10}} &
      \multicolumn{1}{c|}{\multirow{1}{*}{3.84}} &
      \multicolumn{1}{c}{\multirow{1}{*}{\textbf{4.72}}} \\
      \multicolumn{1}{c|}{\multirow{1}{*}{}} &
      \multicolumn{1}{c|}{\multirow{1}{*}{s-IoU}} &
      \multicolumn{1}{c|}{\multirow{1}{*}{--}} &
      \multicolumn{1}{c|}{\multirow{1}{*}{--}} &
      \multicolumn{1}{c|}{\multirow{1}{*}{--}} &
      \multicolumn{1}{c|}{\multirow{1}{*}{4.23}} &
      \multicolumn{1}{c|}{\multirow{1}{*}{9.70}} &
      \multicolumn{1}{c|}{\multirow{1}{*}{--}} &
      \multicolumn{1}{c|}{\multirow{1}{*}{--}} &
      \multicolumn{1}{c|}{\multirow{1}{*}{--}} &
      \multicolumn{1}{c|}{\multirow{1}{*}{--}} &
      \multicolumn{1}{c}{\multirow{1}{*}{\textbf{56.82}}} \\
      \multicolumn{1}{c|}{\multirow{1}{*}{}} &
      \multicolumn{1}{c|}{\multirow{1}{*}{t-IoU}} &
      \multicolumn{1}{c|}{\multirow{1}{*}{--}} &
      \multicolumn{1}{c|}{\multirow{1}{*}{--}} &
      \multicolumn{1}{c|}{\multirow{1}{*}{--}} &
      \multicolumn{1}{c|}{\multirow{1}{*}{6.94}} &
      \multicolumn{1}{c|}{\multirow{1}{*}{11.54}} &
      \multicolumn{1}{c|}{\multirow{1}{*}{7.20}} &
      \multicolumn{1}{c|}{\multirow{1}{*}{12.15}} &
      \multicolumn{1}{c|}{\multirow{1}{*}{34.61}} &
      \multicolumn{1}{c|}{\multirow{1}{*}{39.09}} &
      \multicolumn{1}{c}{\multirow{1}{*}{\textbf{68.35}}} \\
      \hline
      \multicolumn{1}{c|}{\multirow{4}{*}{\makecell{Real\\Event\\-ST}}} &
      \multicolumn{1}{c|}{\multirow{1}{*}{DC}} &
      \multicolumn{1}{c|}{\multirow{1}{*}{1.98}} &
      \multicolumn{1}{c|}{\multirow{1}{*}{1.82}} &
      \multicolumn{1}{c|}{\multirow{1}{*}{2.30}} &
      \multicolumn{1}{c|}{\multirow{1}{*}{2.26}} &
      \multicolumn{1}{c|}{\multirow{1}{*}{2.55}} &
      \multicolumn{1}{c|}{\multirow{1}{*}{2.20}} &
      \multicolumn{1}{c|}{\multirow{1}{*}{2.80}} &
      \multicolumn{1}{c|}{\multirow{1}{*}{2.45}} &
      \multicolumn{1}{c|}{\multirow{1}{*}{2.92}} &
      \multicolumn{1}{c}{\multirow{1}{*}{\textbf{3.65}}} \\
      \multicolumn{1}{c|}{\multirow{1}{*}{}} &
      \multicolumn{1}{c|}{\multirow{1}{*}{VQA}} &
      \multicolumn{1}{c|}{\multirow{1}{*}{2.31}} &
      \multicolumn{1}{c|}{\multirow{1}{*}{1.80}} &
      \multicolumn{1}{c|}{\multirow{1}{*}{2.45}} &
      \multicolumn{1}{c|}{\multirow{1}{*}{3.04}} &
      \multicolumn{1}{c|}{\multirow{1}{*}{3.62}} &
      \multicolumn{1}{c|}{\multirow{1}{*}{3.11}} &
      \multicolumn{1}{c|}{\multirow{1}{*}{3.75}} &
      \multicolumn{1}{c|}{\multirow{1}{*}{3.18}} &
      \multicolumn{1}{c|}{\multirow{1}{*}{3.93}} &
      \multicolumn{1}{c}{\multirow{1}{*}{\textbf{4.82}}} \\
      \multicolumn{1}{c|}{\multirow{1}{*}{}} &
      \multicolumn{1}{c|}{\multirow{1}{*}{s-IoU}} &
      \multicolumn{1}{c|}{\multirow{1}{*}{--}} &
      \multicolumn{1}{c|}{\multirow{1}{*}{--}} &
      \multicolumn{1}{c|}{\multirow{1}{*}{--}} &
      \multicolumn{1}{c|}{\multirow{1}{*}{4.02}} &
      \multicolumn{1}{c|}{\multirow{1}{*}{7.97}} &
      \multicolumn{1}{c|}{\multirow{1}{*}{--}} &
      \multicolumn{1}{c|}{\multirow{1}{*}{--}} &
      \multicolumn{1}{c|}{\multirow{1}{*}{--}} &
      \multicolumn{1}{c|}{\multirow{1}{*}{--}} &
      \multicolumn{1}{c}{\multirow{1}{*}{\textbf{46.27}}} \\
      \multicolumn{1}{c|}{\multirow{1}{*}{}} &
      \multicolumn{1}{c|}{\multirow{1}{*}{t-IoU}} &
      \multicolumn{1}{c|}{\multirow{1}{*}{--}} &
      \multicolumn{1}{c|}{\multirow{1}{*}{--}} &
      \multicolumn{1}{c|}{\multirow{1}{*}{--}} &
      \multicolumn{1}{c|}{\multirow{1}{*}{6.23}} &
      \multicolumn{1}{c|}{\multirow{1}{*}{10.61}} &
      \multicolumn{1}{c|}{\multirow{1}{*}{5.35}} &
      \multicolumn{1}{c|}{\multirow{1}{*}{7.40}} &
      \multicolumn{1}{c|}{\multirow{1}{*}{21.05}} &
      \multicolumn{1}{c|}{\multirow{1}{*}{23.55}} &
      \multicolumn{1}{c}{\multirow{1}{*}{\textbf{52.80}}} \\
       \Xhline{1pt}
  \end{tabular}
  \caption{Quantitative results of LMMs with spatiotemporal understanding on different datasets.} 
   \label{Tab:Comparison}
\end{table*}

\begin{figure*}
  \setlength{\abovecaptionskip}{5pt}
  \setlength{\belowcaptionskip}{-8pt}
  \centering
   \includegraphics[width=0.97\linewidth]{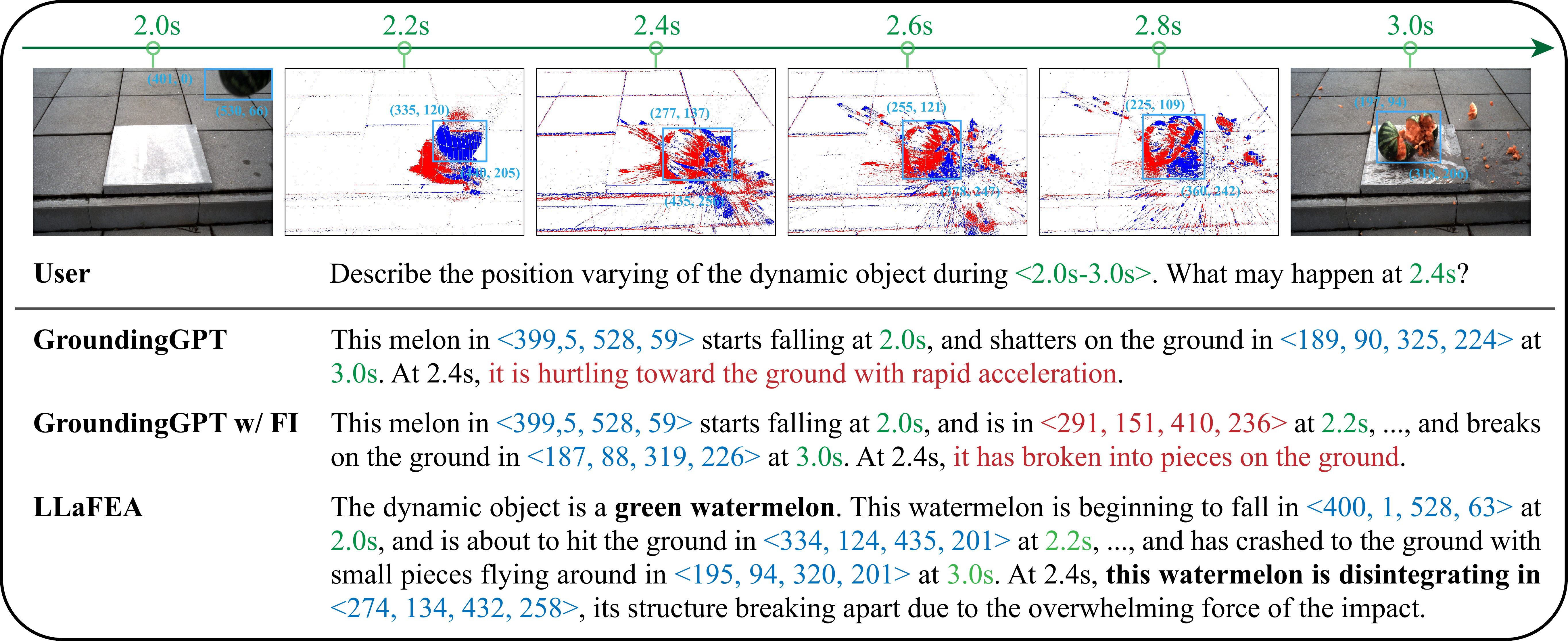}
   \caption{Visual comparison of LMMs on fine-grained understanding in high-dynamic scenes. Red fonts highlight incorrect results.
   }
   \label{Fig:Comparison_Dynamic}
\end{figure*}

\vspace{1mm}
\noindent
\textbf{LLaFEA-RealEvent-ST.} 
\gh{Due to the domain gap between synthetic and real data, LMMs trained solely on synthetic datasets struggle to generalize to complex real-world scenes. We thus incorporate real event datasets: DSEC~\cite{Gehrig21ral} and CoSEC~\cite{peng2024cosec} into a new frame-event dataset with 74K samples that include challenging conditions such as low-light and high-dynamic environments. For data alignment, frames and event data in DSEC are synchronized via stereo matching, and those in CoSEC are physically aligned using a coaxial optical device. To generate text annotations, we employ GPT-4V~\cite{yang2023dawn} to produce understanding texts from multiple images. Additionally, we extract bounding boxes using object detection algorithms~\cite{girshick2015fast} for spatial positions and derive temporal durations from event streams. These elements are integrated into an instruction-following format via GPT-4V for fine-grained spatiotemporal dense understanding.
}

\subsection{Training Pipelines}
To ensure the stability of the training process and improve the performance of the model, we divide the entire training into four stages in Table \ref{Tab:Dataset} as follows:

\vspace{1mm}
\noindent 
\textbf{Stage 1: Language-Vision Content Alignment.}
\gh{Caption data in LLaFEA-SynEvent-ST is used to initially align linguistic and visual representations, providing a foundational understanding for the proposed model. At this stage, only the projector parameters are trained. The spatiotemporal coordinate tokens \( \tilde{v} \) are set to zero.}

\vspace{1mm}
\noindent 
\textbf{Stage 2: Frame-Event Fusion and Matching.}
\gh{We train the frame-event spatiotemporal fusion process using caption and VQA data from LLaFEA-SynEvent-ST for the model to generate spatiotemporal-dense features. At this stage, we update the parameters of cross-attention fusion and self-attention matching while keeping all other modules frozen.}

\vspace{1mm}
\noindent 
\textbf{Stage 3: Spatial-Temporal Coordinate Alignment.}
\gh{
To improve fine-grained spatiotemporal understanding, we use the spatial and temporal grounding data from LLaFEA-SynEvent-ST to refine spatiotemporal coordinate alignment between linguistic and visual representations. At this stage, only the parameters of the coordinate embedding and self-attention fusion modules are updated.}

\vspace{1mm}
\noindent 
\textbf{Stage 4: Instruction Fine-Tuning.}
\gh{Our LLaFEA-RealEvent-ST dataset is used to improve generalization capability of the model for fine-grained spatiotemporal-dense understanding through a multi-task instruction fine-tuning strategy. At this stage, all trainable parameters are updated while the frame-based visual encoder remains frozen.}

\begin{figure*}
  \setlength{\abovecaptionskip}{5pt}
  \setlength{\belowcaptionskip}{-14pt}
  \centering
   \includegraphics[width=0.95\linewidth]{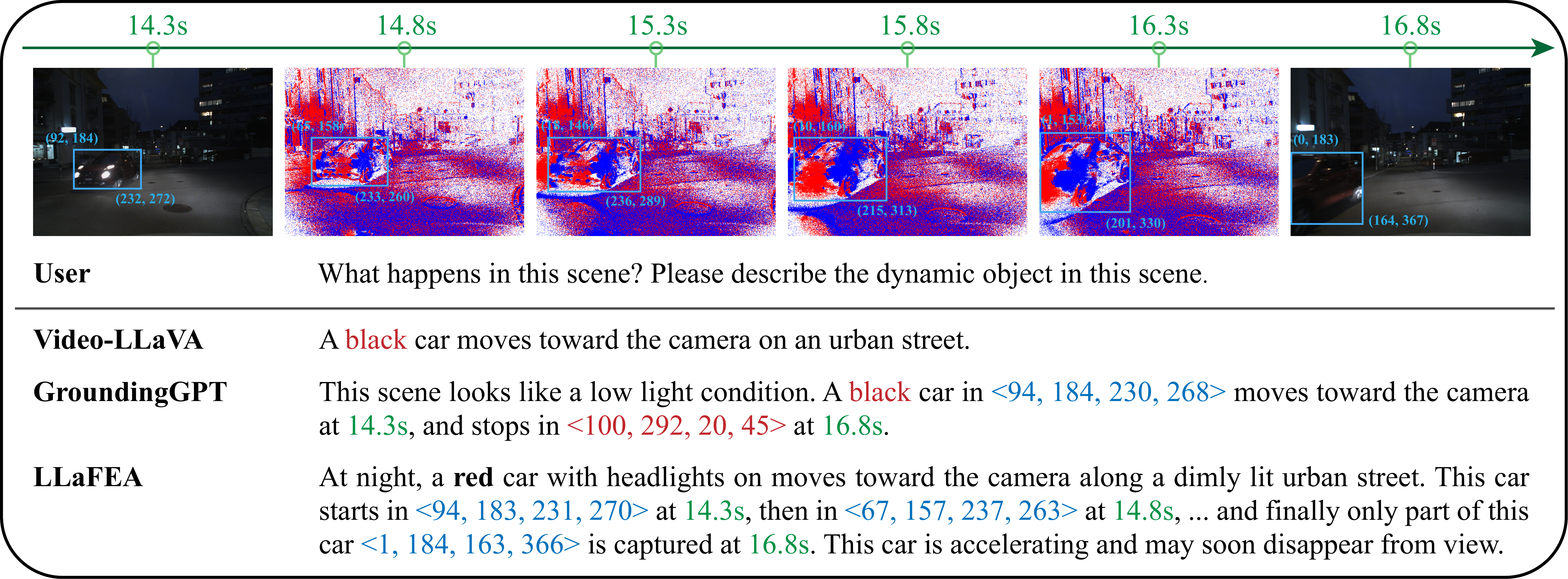}
   \caption{Visual comparison of LMMs on spatiotemporal understanding in low-light scenes. Red fonts highlight incorrect results.
   } 
   \label{Fig:Comparison_LowLight}
\end{figure*}

\section{Experiments}
\label{sec:experiments}
\subsection{Experiment Setup}

\noindent
\textbf{Implements Details.}
\gh{Our LLaFEA model utilizes the pre-trained weights of LLaVA-1.5-7B~\cite{liu2024improved} and the frame-based visual encoder of CLIP-ViT-L-336px~\cite{cherti2023reproducible, radford2021learning}. The event-based visual encoder follows the ViT architecture and is initialized with pre-trained weights~\cite{wu2023eventclip, klenk2024masked} from N-ImageNet~\cite{kim2021n}. Cross-attention fusion and self-attention matching share the same architecture. The model is trained on 8 RTX 4090 GPUs over 68 hours using AdamW as the optimizer. In training stages 1-3, we set the learning rate to \(1.0e-4\) with a batch size of 48. We use a learning rate of \(1.5e-5\) with a batch size of 36 in stage 4.}


\vspace{1mm}
\noindent
\textbf{Comparison Methods.}
\gh{We compare our approach with LMMs capable of processing video inputs: Video-ChatGPT~\cite{maaz2023video}, Video-LLaVA~\cite{lin2023video}, GroundingGPT~\cite{li2024groundinggpt}, Grounded-VideoLLM~\cite{wang2024grounded}, and VTimeLLM~\cite{huang2024vtimellm}. For a fair comparison, all methods use a 7B large language model and are trained on the same instruction fine-tuning dataset.
We employ two training strategies for the competing methods: 1) Directly training LMMs on the dataset; 2) Using event-based frame interpolation~\cite{tulyakov2021time} (denoted as "w/ FI") to generate temporally dense videos for LMM training. The latter strategy standardizes the modality type of visual representation across all methods for fair experimental comparisons.}

\begin{table}\footnotesize
    \setlength{\abovecaptionskip}{5pt}
    \setlength\tabcolsep{4pt}
    \setlength{\belowcaptionskip}{-8pt}
  \centering
  \renewcommand\arraystretch{1.05}
  \begin{tabular}{ccccccc}
    \Xhline{1pt}
      \multicolumn{2}{c|}{\multirow{1}{*}{Cross-Attention}} &
      \multicolumn{1}{c|}{\multirow{2}{*}{\makecell{Self-Attention\\Matching}}} & \multicolumn{1}{c}{\multirow{2}{*}{DC}} & \multicolumn{1}{c}{\multirow{2}{*}{VQA}} & \multicolumn{1}{c}{\multirow{2}{*}{s-IoU}} & \multicolumn{1}{c}{\multirow{2}{*}{t-IoU}}\\
      \cline{1-2}
      \multicolumn{1}{c|}{\multirow{1}{*}{Spatial}} & \multicolumn{1}{c|}{\multirow{1}{*}{Temporal}} & \multicolumn{1}{c|}{\multirow{1}{*}{}}\\
      \hline 
      \multicolumn{1}{c|}{\multirow{1}{*}{$\times$}} & 
      \multicolumn{1}{c|}{\multirow{1}{*}{$\times$}} & 
      \multicolumn{1}{c|}{\multirow{1}{*}{$\times$}} & 2.25 & 3.10 & 4.03 & 6.15\\

      \multicolumn{1}{c|}{\multirow{1}{*}{$\surd$}} & 
      \multicolumn{1}{c|}{\multirow{1}{*}{$\times$}} & 
      \multicolumn{1}{c|}{\multirow{1}{*}{$\times$}} & 2.39 & 3.27 & 15.40 & 6.16\\

      \multicolumn{1}{c|}{\multirow{1}{*}{$\times$}} & 
      \multicolumn{1}{c|}{\multirow{1}{*}{$\surd$}} & 
      \multicolumn{1}{c|}{\multirow{1}{*}{$\times$}} & 2.30 & 3.25 & 4.53 & 22.74\\

      \multicolumn{1}{c|}{\multirow{1}{*}{$\times$}} & 
      \multicolumn{1}{c|}{\multirow{1}{*}{$\times$}} & 
      \multicolumn{1}{c|}{\multirow{1}{*}{$\surd$}} & 3.08 & 4.10 & 12.35 & 21.02\\

      \multicolumn{1}{c|}{\multirow{1}{*}{$\surd$}} & 
      \multicolumn{1}{c|}{\multirow{1}{*}{$\surd$}} & 
      \multicolumn{1}{c|}{\multirow{1}{*}{$\surd$}} & \textbf{3.65} & \textbf{4.82} & \textbf{46.27} & \textbf{52.80} \\

       \Xhline{1pt}
  \end{tabular}
  \caption{Effectiveness of hierarchical fusion framework.} 
   \label{Tab:Ablation_Framework}
\end{table}

\vspace{1mm}
\noindent
\textbf{Evaluation Metric.}
\gh{To assess both coarse- and fine-grained spatiotemporal understanding, we evaluate on three key tasks: detailed captioning (DC), visual question answering (VQA), and spatiotemporal grounding (STG), with STG specifically reflecting fine-grained spatiotemporal comprehension.
For DC, we measure the completeness of key detail descriptions. For VQA, we assess the precision of textual understanding in answering specific questions. For STG, we use spatial and temporal intersection over union (s-IoU/t-IoU) as evaluation metrics.}

\subsection{Comparison with State-of-the-Art Models}

\noindent
\textbf{Quantitative Results.}
\gh{In Table~\ref{Tab:Comparison}, we compare competing methods using the two training strategies. 1) Frame-event methods consistently outperform frame-only methods showing that event-based data effectively compensate for the temporal limitations of frame cameras to improve spatiotemporal understanding of LMMs. 2) Our proposed method achieves the best performance in the spatiotemporal grounding task, outperforming other frame-event methods. This is because our approach captures global spatiotemporal associations for fine-grained understanding. In contrast, competing methods simply fuse frame-event data at the visual input level without explicit spatiotemporal matching.
}

\begin{table}\footnotesize
    \setlength{\abovecaptionskip}{5pt}
    \setlength\tabcolsep{10pt}
    \setlength{\belowcaptionskip}{-12pt}
  \centering
  \renewcommand\arraystretch{1.05}
  \begin{tabular}{ccccc}
    \Xhline{1pt}
      \multicolumn{1}{c|}{\multirow{1}{*}{Vision input}} &
      \multicolumn{1}{c}{\multirow{1}{*}{DC}} & \multicolumn{1}{c}{\multirow{1}{*}{VQA}} & \multicolumn{1}{c}{\multirow{1}{*}{s-IoU}} & \multicolumn{1}{c}{\multirow{1}{*}{t-IoU}}\\
      \hline
      \multicolumn{1}{c|}{\multirow{1}{*}{w/ Frame}} &
      \multicolumn{1}{c}{\multirow{1}{*}{3.61}} & \multicolumn{1}{c}{\multirow{1}{*}{4.74}} & \multicolumn{1}{c}{\multirow{1}{*}{40.52}} & \multicolumn{1}{c}{\multirow{1}{*}{22.05}}\\
      
      \multicolumn{1}{c|}{\multirow{1}{*}{w/ Event}} &
      \multicolumn{1}{c}{\multirow{1}{*}{3.16}} & \multicolumn{1}{c}{\multirow{1}{*}{4.30}} & \multicolumn{1}{c}{\multirow{1}{*}{35.75}} & \multicolumn{1}{c}{\multirow{1}{*}{50.28}}\\
      
      \multicolumn{1}{c|}{\multirow{1}{*}{w/ Frame-Event}} &
      \multicolumn{1}{c}{\multirow{1}{*}{\textbf{3.65}}} & \multicolumn{1}{c}{\multirow{1}{*}{\textbf{4.82}}} & \multicolumn{1}{c}{\multirow{1}{*}{\textbf{46.27}}} & \multicolumn{1}{c}{\multirow{1}{*}{\textbf{52.80}}}\\
       \Xhline{1pt}
  \end{tabular}
  \caption{Effect of vision input on spatiotemporal understanding.} 
   \label{Tab:Ablation_VisionInput}
\end{table}

\vspace{1mm}
\noindent
\textbf{Qualitative Results.}
\gh{We evaluate fine-grained spatiotemporal understanding under high-dynamic and low-light conditions, which serve as representative challenging scenes.
For high-dynamic scenes (Fig.~\ref{Fig:Comparison_Dynamic}), frame-only methods struggle with temporal sparsity, capturing object positions at only a few timestamps. Although frame-event methods can track position changes over time, they suffer from spatial misalignment due to the absence of spatiotemporal matching. In contrast, our method accurately infers continuous spatial position changes over time. For low-light scenes (Fig.~\ref{Fig:Comparison_LowLight}), competing methods generate largely incorrect descriptions while our approach accurately interprets the scene. This highlights how the high dynamic range of event data is effectively integrated into spatiotemporal feature fusion for robust scene understanding under extreme illumination conditions.
These results demonstrate that the use of event cameras significantly improves the generalization ability of our method on challenging real-world scenarios.
}

\subsection{Ablation Study}
\noindent
\textbf{How Hierarchical Fusion Framework works?}
\gh{In Table~\ref{Tab:Ablation_Framework}, we validate the effectiveness of the proposed hierarchical fusion framework for integrating frame and event modalities. The results show that cross-attention spatial fusion impacts the spatial grounding metric, and cross-attention temporal fusion influences the temporal grounding metric. Additionally, self-attention matching improves both spatiotemporal understanding metrics. These findings indicate that spatiotemporal matching establishes the fundamental spatiotemporal understanding capability, and spatiotemporal fusion further enhances fine-grained spatiotemporal reasoning.}

\noindent
\textbf{Influence of Frame and Event Modalities.}
\gh{Table~\ref{Tab:Ablation_VisionInput} studies the impact of frame and event modalities on spatiotemporal understanding. Our method performs well on DC, VQA, and spatial grounding using only the frame modality, but struggles with temporal grounding due to temporal sparsity. Using only the event modality improves temporal grounding, but degrades performance in DC, VQA, and spatial grounding due to the lack of spatial detail.
Our method achieves a significant improvement in fine-grained spatiotemporal understanding when both modalities are fused since frame-event fusion enhances the spatiotemporal dense correspondence between linguistic and visual representations.}

\subsection{Discussion}
\begin{figure}
  \setlength{\abovecaptionskip}{4pt}
  \setlength{\belowcaptionskip}{-12pt}
  \centering
   \includegraphics[width=0.97\linewidth]{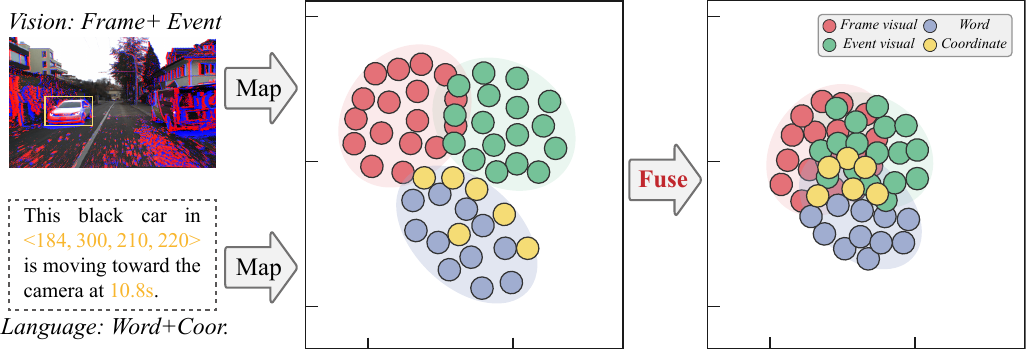}
   \caption{
   \gh{t-SNE visualization of feature tokens. Frame-event spatiotemporal fusion aligns visual and linguistic tokens, with coordinate tokens acting as a bridge between them.}
   }
   \label{Fig:Discussion_FusionforCoor}
\end{figure}

\noindent
\textbf{Role of Spatiotemporal Fusion on Coordinate Alignment.}
\gh{Fig.~\ref{Fig:Discussion_FusionforCoor} visualizes feature tokens to show the impact of spatiotemporal fusion on coordinate alignment. The results show that frame-event spatiotemporal fusion facilitates the coordinate token in acting as an intermediate bridge to align linguistic and visual representation tokens. This alignment of language-vision spatiotemporal dense correspondences enhances fine-grained spatiotemporal understanding.}

\noindent
\textbf{Impact of Spatiotemporal Feature Fusion Strategy.}
\gh{Table~\ref{Tab:Discussion_FusionStrategy} compares the impact of different spatiotemporal fusion strategies: concatenation, weighting, and attention on spatiotemporal understanding. The results show that attention-based fusion outperforms other methods. This is because concatenation and weighting rely on global unified fusion with fixed weights. Attention-based fusion provides dynamic adaptability and local perception for more effective modeling of inter-modal spatiotemporal relationships.}

\noindent
\textbf{Importance of Coordinate Embedding.}
\gh{We examine the role of spatiotemporal coordinate embedding in fine-grained spatiotemporal understanding. Table~\ref{Tab:Discussion_CoorAlign} shows that coordinate embedding significantly increases fine-grained performance despite
having little impact on coarse-grained spatiotemporal understanding. This improvement comes from its ability to reinforce pixel-level and object-level spatiotemporal correspondence between linguistic and visual representations.}

\begin{table}\footnotesize
    \setlength{\abovecaptionskip}{4pt}
    \setlength\tabcolsep{10pt}
    \setlength{\belowcaptionskip}{-8pt}
  \centering
  \renewcommand\arraystretch{1.05}
  \begin{tabular}{ccccc}
    \Xhline{1pt}
      \multicolumn{1}{c|}{\multirow{1}{*}{Fusion strategy}} &
      \multicolumn{1}{c}{\multirow{1}{*}{DC}} & \multicolumn{1}{c}{\multirow{1}{*}{VQA}} & \multicolumn{1}{c}{\multirow{1}{*}{s-IoU}} & \multicolumn{1}{c}{\multirow{1}{*}{t-IoU}}\\
      \hline
      \multicolumn{1}{c|}{\multirow{1}{*}{w/ Concatenation}} &
      \multicolumn{1}{c}{\multirow{1}{*}{3.05}} & \multicolumn{1}{c}{\multirow{1}{*}{3.88}} & \multicolumn{1}{c}{\multirow{1}{*}{35.70}} & \multicolumn{1}{c}{\multirow{1}{*}{40.32}}\\
      
      \multicolumn{1}{c|}{\multirow{1}{*}{w/ Weighting}} &
      \multicolumn{1}{c}{\multirow{1}{*}{3.09}} & \multicolumn{1}{c}{\multirow{1}{*}{3.90}} & \multicolumn{1}{c}{\multirow{1}{*}{36.85}} & \multicolumn{1}{c}{\multirow{1}{*}{41.74}}\\

      \multicolumn{1}{c|}{\multirow{1}{*}{w/ Attention}} &
      \multicolumn{1}{c}{\multirow{1}{*}{\textbf{3.65}}} & \multicolumn{1}{c}{\multirow{1}{*}{\textbf{4.82}}} & \multicolumn{1}{c}{\multirow{1}{*}{\textbf{46.27}}} & \multicolumn{1}{c}{\multirow{1}{*}{\textbf{52.80}}}\\

       \Xhline{1pt}
  \end{tabular}
  \caption{Choice of spatiotemporal fusion strategy.} 
   \label{Tab:Discussion_FusionStrategy}
\end{table}

\begin{table}\footnotesize
    \setlength{\abovecaptionskip}{4pt}
    \setlength\tabcolsep{10pt}
    \setlength{\belowcaptionskip}{-8pt}
  \centering
  \renewcommand\arraystretch{1.05}
  \begin{tabular}{ccccc}
    \Xhline{1pt}
      \multicolumn{1}{c|}{\multirow{1}{*}{Coordinate align}} &
      \multicolumn{1}{c}{\multirow{1}{*}{DC}} & \multicolumn{1}{c}{\multirow{1}{*}{VQA}} & \multicolumn{1}{c}{\multirow{1}{*}{s-IoU}} & \multicolumn{1}{c}{\multirow{1}{*}{t-IoU}}\\
      \hline
      \multicolumn{1}{c|}{\multirow{1}{*}{w/o Embedding}} &
      \multicolumn{1}{c}{\multirow{1}{*}{3.62}} & \multicolumn{1}{c}{\multirow{1}{*}{4.80}} & \multicolumn{1}{c}{\multirow{1}{*}{14.15}} & \multicolumn{1}{c}{\multirow{1}{*}{18.36}}\\
      
      \multicolumn{1}{c|}{\multirow{1}{*}{w/ Embedding}} &
      \multicolumn{1}{c}{\multirow{1}{*}{\textbf{3.65}}} & \multicolumn{1}{c}{\multirow{1}{*}{\textbf{4.82}}} & \multicolumn{1}{c}{\multirow{1}{*}{\textbf{46.27}}} & \multicolumn{1}{c}{\multirow{1}{*}{\textbf{52.80}}}\\

       \Xhline{1pt}
  \end{tabular}
  \caption{Discussion on coordinate embedding.}
   \label{Tab:Discussion_CoorAlign}
\end{table}

\noindent
\textbf{Joint Training \emph{v.s.} Separate Training.}
In Table \ref{Tab:Discussion_TrainingPipeline}, we study the effects of different training pipelines on the proposed model. We 
observe that joint training takes more time and performs worse than separate training. This is because the entire framework contains multiple unimodal models: 
frame/event-based encoders and LLM, and \gh{thus} the direct joint training may cause mutual interference between different models, making it difficult to ensure the alignment of cross-modal spatiotemporal features.

\begin{table}\footnotesize
    \setlength{\abovecaptionskip}{4pt}
    \setlength\tabcolsep{6pt}
    \setlength{\belowcaptionskip}{-11pt}
  \centering
  \renewcommand\arraystretch{1.05}
  \begin{tabular}{cccccc}
    \Xhline{1pt}
      \multicolumn{1}{c|}{\multirow{1}{*}{Training pipeline}} &
      \multicolumn{1}{c}{\multirow{1}{*}{Time}} &
      \multicolumn{1}{c}{\multirow{1}{*}{DC}} & \multicolumn{1}{c}{\multirow{1}{*}{VQA}} & \multicolumn{1}{c}{\multirow{1}{*}{s-IoU}} & \multicolumn{1}{c}{\multirow{1}{*}{t-IoU}}\\
      \hline
      \multicolumn{1}{c|}{\multirow{1}{*}{w/ Joint training}} &
      \multicolumn{1}{c}{\multirow{1}{*}{94 h}} & \multicolumn{1}{c}{\multirow{1}{*}{3.28}} & \multicolumn{1}{c}{\multirow{1}{*}{4.36}} & \multicolumn{1}{c}{\multirow{1}{*}{38.05}} & \multicolumn{1}{c}{\multirow{1}{*}{47.54}}\\
      
      \multicolumn{1}{c|}{\multirow{1}{*}{w/ Separate training}} &
      \multicolumn{1}{c}{\multirow{1}{*}{\textbf{68 h}}} & \multicolumn{1}{c}{\multirow{1}{*}{\textbf{3.65}}} & \multicolumn{1}{c}{\multirow{1}{*}{\textbf{4.82}}} & \multicolumn{1}{c}{\multirow{1}{*}{\textbf{46.27}}} & \multicolumn{1}{c}{\multirow{1}{*}{\textbf{52.80}}}\\

       \Xhline{1pt}
  \end{tabular}
  \caption{Comparison of different training pipelines.} 
   \label{Tab:Discussion_TrainingPipeline}
\end{table}

\noindent
\textbf{Limitation.}
\gh{Our method performs well in most scenes but struggles with smooth, single-colored environments (\emph{e.g.}, white wall backgrounds). Since event cameras trigger signals based on brightness changes, these textures make it difficult to capture valid events, leading to abnormal spatiotemporal fusion and matching. Our future work will incorporate depth sensors (\emph{e.g.}, LiDAR and RGB-D) to enhance structural modeling and improve scene understanding in LMMs.}

\section{Conclusion}
\label{sec:conclusion}
\gh{We introduce an event-frame fusion approach and propose a large language and frame-event model for scene understanding. Our hierarchical fusion framework integrates spatial-dense, temporal-sparse frame features with spatial-sparse, temporal-dense event features to generate spatiotemporal-dense visual representations. We further embed textual spatiotemporal coordinates into the fused visual features for LMMs to achieve fine-grained spatiotemporal understanding. Additionally, we construct a real frame-event dataset with spatiotemporal coordinate instructions for fine-tuning and conduct extensive experiments to validate the effectiveness of our method.}

{
  \small
  \bibliographystyle{ieeenat_fullname}
  \bibliography{egbib}

\begin{thebibliography}{10}

\bibitem{alayrac2022flamingo}
Jean-Baptiste Alayrac, Jeff Donahue, Pauline Luc, Antoine Miech, Iain Barr, Yana Hasson, Arthur Mensch, Katie Millican, David Moore, Michael Needham, et~al.
\newblock Flamingo: a visual language model for few-shot learning.
\newblock {\em Advances in Neural Information Processing Systems}, 35:23716--23732, 2022.

\bibitem{li2022blip}
Junnan Li, Dongxu Li, Caiming Xiong, and Steven Hoi.
\newblock {BLIP}: Bootstrapping language-image pre-training for unified vision-language understanding and generation.
\newblock In {\em Proceedings of the 39th International Conference on Machine Learning}, volume 162 of {\em Proceedings of Machine Learning Research}, pages 12888--12900. PMLR, 2022.

\bibitem{li2023blip2}
Junnan Li, Dongxu Li, Silvio Savarese, and Steven Hoi.
\newblock {BLIP}-2: Bootstrapping language-image pre-training with frozen image encoders and large language models.
\newblock In {\em Proceedings of the 40th International Conference on Machine Learning}, volume 202 of {\em Proceedings of Machine Learning Research}, pages 19730--19742. PMLR, 2023.

\bibitem{liu2023visual}
Haotian Liu, Chunyuan Li, Qingyang Wu, and Yong~Jae Lee.
\newblock Visual instruction tuning.
\newblock In {\em Advances in Neural Information Processing Systems (NeurIPS)}, 2023.

\bibitem{radford2021learning}
Alec Radford, Jong~Wook Kim, Chris Hallacy, Aditya Ramesh, Gabriel Goh, Sandhini Agarwal, Girish Sastry, Amanda Askell, Pam Mishkin, Jack Clark, et~al.
\newblock Learning transferable visual models from natural language supervision.
\newblock In {\em International Conference on Machine Learning}, pages 8748--8763. PMLR, 2021.

\bibitem{gong2022contrastive}
Yujia Gong, Changhan Wang, Yun~Tang Wang, and Jiatao Gu.
\newblock Contrastive learning for speech translation.
\newblock In {\em International Conference on Learning Representations}, 2022.

\bibitem{wang2022git}
Jianfeng Wang, Jianwei Yang, Xiaowei Wang, Lu~Yuan, Lei Zhang, Yejin Choi, and Jianfeng Gao.
\newblock Git: A generative image-to-text transformer for vision and language.
\newblock In {\em Advances in Neural Information Processing Systems}, volume~35, pages 18002--18014, 2022.

\bibitem{kamath2021mdetr}
Aishwarya Kamath, Mannat Singh, Yann LeCun, Ishan Misra, Gabriel Synnaeve, Nicolas Carion, and Karteek Alahari.
\newblock Mdetr: Modulated detection for end-to-end multi-modal understanding.
\newblock In {\em Proceedings of the IEEE/CVF International Conference on Computer Vision}, pages 1780--1790, 2021.

\bibitem{liu2023llava}
Haotian Liu, Chunyuan Li, Qingyang Wu, and Yong~Jae Lee.
\newblock Llava: Large language and vision assistant.
\newblock {\em arXiv preprint arXiv:2304.08485}, 2023.

\bibitem{chen2023pali}
Mandy Chen, Adams~Wei Yu, Hamid Palangi, Paul Smolensky, Yinfei Yang, Xiaowei Yuan, Kathy Meier-Hellstern, Jianfeng Gao, Ed~Chi, et~al.
\newblock Pali: A jointly-scaled multilingual language-image model.
\newblock {\em arXiv preprint arXiv:2303.07892}, 2023.

\bibitem{chen2017deeplab}
Liang{-}Chieh Chen, George Papandreou, Iasonas Kokkinos, Kevin Murphy, and Alan~L. Yuille.
\newblock Deeplab: Semantic image segmentation with deep convolutional nets, atrous convolution, and fully connected crfs.
\newblock {\em IEEE Transactions on Pattern Analysis and Machine Intelligence}, 40(4):834--848, 2018.

\bibitem{eigen2014depth}
David Eigen, Christian Puhrsch, and Rob Fergus.
\newblock Depth map prediction from a single image using a multi-scale deep network.
\newblock In {\em Advances in Neural Information Processing Systems (NeurIPS)}, pages 2366--2374, 2014.

\bibitem{dosovitskiy2015flownet}
Alexey Dosovitskiy, Philipp Fischer, Eddy Ilg, Philip Hausser, Caner Hazirbas, Vladimir Golkov, Patrick van~der Smagt, Daniel Cremers, and Thomas Brox.
\newblock Flownet: Learning optical flow with convolutional networks.
\newblock In {\em IEEE International Conference on Computer Vision (ICCV)}, pages 2758--2766, 2015.

\bibitem{brown2020gpt3}
Tom~B. Brown, Benjamin Mann, Nick Ryder, Melanie Subbiah, Jared Kaplan, Prafulla Dhariwal, Arvind Neelakantan, Pranav Shyam, Girish Sastry, Amanda Askell, et~al.
\newblock Language models are few-shot learners.
\newblock {\em Advances in Neural Information Processing Systems (NeurIPS)}, 33:1877--1901, 2020.

\bibitem{liu2023improved}
Haotian Liu, Pengchuan Zhang, Jianwei Yang, and Lei Zhang.
\newblock Improved visual instruction tuning for image-text alignment.
\newblock {\em arXiv preprint}, abs/2310.00067, 2023.

\bibitem{lei2021momentdetr}
Jie Lei, Licheng Yu, Tamara~L. Berg, and Mohit Bansal.
\newblock Less is more: Clipbert for video-and-language learning via sparse sampling.
\newblock In {\em IEEE Conference on Computer Vision and Pattern Recognition (CVPR)}, pages 7331--7341, 2021.

\bibitem{xu2021videoclip}
Mingda Xu, Yannis Kalantidis, Marcus Rohrbach, Kristen Grauman, and Trevor Darrell.
\newblock Videoclip: Contrastive pretraining for zero-shot video-text understanding.
\newblock {\em Advances in Neural Information Processing Systems (NeurIPS)}, 34:6606--6620, 2021.

\bibitem{luo2022clip4clip}
Lei Luo, Xun Guo, Qian Wu, and Zhibo Chen.
\newblock Clip4clip: An empirical study of clip for end-to-end video clip retrieval and captioning.
\newblock {\em arXiv preprint}, abs/2203.00551, 2022.

\bibitem{akbari2021vatt}
Hassan Akbari, Linagzhe Zhu, Joseph Chang, Ivan Krasin, George Toderici, Yonghui Wu, and Cordelia Schmid.
\newblock Vatt: Transformers for multimodal self-supervised learning from raw video, audio and text.
\newblock {\em Advances in Neural Information Processing Systems (NeurIPS)}, 34:24206--24221, 2021.

\bibitem{gallego2020event}
Guillermo Gallego, Tobi Delbruck, Garrick Orchard, Chiara Bartolozzi, Brian Taba, Andrea Censi, et~al.
\newblock Event-based vision: A survey.
\newblock {\em IEEE Transactions on Pattern Analysis and Machine Intelligence}, 44(1):154--180, 2020.

\bibitem{posch2014retinomorphic}
Christoph Posch, Daniel Matolin, and Rainer Wohlgenannt.
\newblock A retinomorphic approach to high-speed and high-dynamic-range vision using neuromorphic imaging sensors.
\newblock {\em IEEE Transactions on Circuits and Systems I: Regular Papers}, 61(2):478--500, 2014.

\bibitem{tulyakov2018softmax}
Sergey Tulyakov, Anton Ivanov, and Francois Fleuret.
\newblock Softmax splitting for video frame interpolation.
\newblock In {\em IEEE Conference on Computer Vision and Pattern Recognition (CVPR)}, pages 5437--5446, 2018.

\bibitem{shedligeri2019photorealistic}
Parag~Shiledar Shedligeri and Kaushik Mitra.
\newblock Photorealistic video interpolation using event cameras.
\newblock In {\em IEEE International Conference on Computer Vision (ICCV)}, pages 5172--5181, 2019.

\bibitem{scheerlinck2020fast}
Cedric Scheerlinck, Nick Barnes, Ryan~M. Dunn, Xin Yu, and Ronald~G. Drummond.
\newblock Fast image reconstruction with an event camera.
\newblock In {\em IEEE Conference on Computer Vision and Pattern Recognition (CVPR)}, pages 6741--6750, 2020.

\bibitem{rebecq2019e2vid}
Henri Rebecq, Rene Ranftl, Vladlen Koltun, and Davide Scaramuzza.
\newblock High speed and high dynamic range video with an event camera.
\newblock {\em IEEE Transactions on Pattern Analysis and Machine Intelligence}, 43(6):1964--1980, 2019.

\bibitem{gu2022evt}
Yu~Gu, Bojian Wu, Dong Wang, Xianbin Cao, and Guangming Shi.
\newblock Evt: Event transformers for video recognition.
\newblock In {\em European Conference on Computer Vision (ECCV)}, pages 299--316, 2022.

\bibitem{zhang2023est}
Xiaotong Zhang, Hongyang Xue, Guoqing Liu, Zhiwei Lin, and Chenchen Zhang.
\newblock Est: Event-stream transformer for asynchronous visual recognition.
\newblock In {\em Neural Information Processing Systems (NeurIPS)}, 2023.

\bibitem{wang2023eformer}
Hao Wang, Zheng Ge, Zhen Xu, and Weiyao Lin.
\newblock Eformer: A transformer-based fusion model for event and frame representations.
\newblock In {\em IEEE Conference on Computer Vision and Pattern Recognition (CVPR)}, pages 12345--12355, 2023.

\bibitem{xu2023estformer}
Shuo Xu, Zheng Zhang, Wei Li, and Yao Wang.
\newblock Est-former: Event-frame spatiotemporal transformer for neuromorphic vision.
\newblock In {\em IEEE International Conference on Computer Vision (ICCV)}, 2023.

\bibitem{vanDerMaaten2008tsne}
Laurens van~der Maaten and Geoffrey Hinton.
\newblock Visualizing data using t-sne.
\newblock {\em Journal of Machine Learning Research}, 9:2579--2605, 2008.

\bibitem{bain2021frozen}
Max Bain, Arsha Nagrani, G{\"u}l Varol, and Andrew Zisserman.
\newblock Frozen in time: A joint video and image encoder for end-to-end retrieval.
\newblock In {\em Proceedings of the IEEE/CVF international conference on computer vision}, pages 1728--1738, 2021.

\bibitem{chen2024panda}
Tsai-Shien Chen, Aliaksandr Siarohin, Willi Menapace, Ekaterina Deyneka, Hsiang-wei Chao, Byung~Eun Jeon, Yuwei Fang, Hsin-Ying Lee, Jian Ren, Ming-Hsuan Yang, et~al.
\newblock Panda-70m: Captioning 70m videos with multiple cross-modality teachers.
\newblock In {\em Proceedings of the IEEE/CVF Conference on Computer Vision and Pattern Recognition}, pages 13320--13331, 2024.

\bibitem{wang2023internvid}
Yi~Wang, Yinan He, Yizhuo Li, Kunchang Li, Jiashuo Yu, Xin Ma, Xinhao Li, Guo Chen, Xinyuan Chen, Yaohui Wang, et~al.
\newblock Internvid: A large-scale video-text dataset for multimodal understanding and generation.
\newblock {\em arXiv preprint arXiv:2307.06942}, 2023.

\bibitem{wang2024internvideo2}
Yi~Wang, Kunchang Li, Xinhao Li, Jiashuo Yu, Yinan He, Guo Chen, Baoqi Pei, Rongkun Zheng, Zun Wang, Yansong Shi, et~al.
\newblock Internvideo2: Scaling foundation models for multimodal video understanding.
\newblock In {\em European Conference on Computer Vision}, pages 396--416. Springer, 2024.

\bibitem{luo2023valley}
Ruipu Luo, Ziwang Zhao, Min Yang, Junwei Dong, Da~Li, Pengcheng Lu, Tao Wang, Linmei Hu, Minghui Qiu, and Zhongyu Wei.
\newblock Valley: Video assistant with large language model enhanced ability.
\newblock {\em arXiv preprint arXiv:2306.07207}, 2023.

\bibitem{maaz2023video}
Muhammad Maaz, Hanoona Rasheed, Salman Khan, and Fahad~Shahbaz Khan.
\newblock Video-chatgpt: Towards detailed video understanding via large vision and language models.
\newblock {\em arXiv preprint arXiv:2306.05424}, 2023.

\bibitem{hu2021v2e}
Yuhuang Hu, Shih-Chii Liu, and Tobi Delbruck.
\newblock v2e: From video frames to realistic dvs events.
\newblock In {\em IEEE Conf. Comput. Vis. Pattern Recog. Worksh.}, pages 1312--1321, 2021.

\bibitem{Gehrig21ral}
Mathias Gehrig, Willem Aarents, Daniel Gehrig, and Davide Scaramuzza.
\newblock {DSEC}: A stereo event camera dataset for driving scenarios.
\newblock {\em IEEE Robotics and Automation Letters}, 2021.

\bibitem{peng2024cosec}
Shihan Peng, Hanyu Zhou, Hao Dong, Zhiwei Shi, Haoyue Liu, Yuxing Duan, Yi~Chang, and Luxin Yan.
\newblock Cosec: A coaxial stereo event camera dataset for autonomous driving.
\newblock {\em arXiv preprint arXiv:2408.08500}, 2024.

\bibitem{yang2023dawn}
Zhengyuan Yang, Linjie Li, Kevin Lin, Jianfeng Wang, Chung-Ching Lin, Zicheng Liu, and Lijuan Wang.
\newblock The dawn of lmms: Preliminary explorations with gpt-4v (ision).
\newblock {\em arXiv preprint arXiv:2309.17421}, 9(1):1, 2023.

\bibitem{girshick2015fast}
Ross Girshick.
\newblock Fast r-cnn.
\newblock In {\em Proceedings of the IEEE international conference on computer vision}, pages 1440--1448, 2015.

\bibitem{liu2024improved}
Haotian Liu, Chunyuan Li, Yuheng Li, and Yong~Jae Lee.
\newblock Improved baselines with visual instruction tuning.
\newblock In {\em Proceedings of the IEEE/CVF Conference on Computer Vision and Pattern Recognition}, pages 26296--26306, 2024.

\bibitem{cherti2023reproducible}
Mehdi Cherti, Romain Beaumont, Ross Wightman, Mitchell Wortsman, Gabriel Ilharco, Cade Gordon, Christoph Schuhmann, Ludwig Schmidt, and Jenia Jitsev.
\newblock Reproducible scaling laws for contrastive language-image learning.
\newblock In {\em Proceedings of the IEEE/CVF conference on computer vision and pattern recognition}, pages 2818--2829, 2023.

\bibitem{wu2023eventclip}
Ziyi Wu, Xudong Liu, and Igor Gilitschenski.
\newblock Eventclip: Adapting clip for event-based object recognition.
\newblock {\em arXiv preprint arXiv:2306.06354}, 2023.

\bibitem{klenk2024masked}
Simon Klenk, David Bonello, Lukas Koestler, Nikita Araslanov, and Daniel Cremers.
\newblock Masked event modeling: Self-supervised pretraining for event cameras.
\newblock In {\em Proceedings of the IEEE/CVF Winter Conference on Applications of Computer Vision}, pages 2378--2388, 2024.

\bibitem{kim2021n}
Junho Kim, Jaehyeok Bae, Gangin Park, Dongsu Zhang, and Young~Min Kim.
\newblock N-imagenet: Towards robust, fine-grained object recognition with event cameras.
\newblock In {\em Proceedings of the IEEE/CVF international conference on computer vision}, pages 2146--2156, 2021.

\bibitem{lin2023video}
Bin Lin, Yang Ye, Bin Zhu, Jiaxi Cui, Munan Ning, Peng Jin, and Li~Yuan.
\newblock Video-llava: Learning united visual representation by alignment before projection.
\newblock {\em arXiv preprint arXiv:2311.10122}, 2023.

\bibitem{li2024groundinggpt}
Zhaowei Li, Qi~Xu, Dong Zhang, Hang Song, Yiqing Cai, Qi~Qi, Ran Zhou, Junting Pan, Zefeng Li, Van~Tu Vu, et~al.
\newblock Groundinggpt: Language enhanced multi-modal grounding model.
\newblock {\em arXiv preprint arXiv:2401.06071}, 2024.

\bibitem{wang2024grounded}
Haibo Wang, Zhiyang Xu, Yu~Cheng, Shizhe Diao, Yufan Zhou, Yixin Cao, Qifan Wang, Weifeng Ge, and Lifu Huang.
\newblock Grounded-videollm: Sharpening fine-grained temporal grounding in video large language models.
\newblock {\em arXiv preprint arXiv:2410.03290}, 2024.

\bibitem{huang2024vtimellm}
Bin Huang, Xin Wang, Hong Chen, Zihan Song, and Wenwu Zhu.
\newblock Vtimellm: Empower llm to grasp video moments.
\newblock In {\em Proceedings of the IEEE/CVF Conference on Computer Vision and Pattern Recognition}, pages 14271--14280, 2024.

\bibitem{tulyakov2021time}
Stepan Tulyakov, Daniel Gehrig, Stamatios Georgoulis, Julius Erbach, Mathias Gehrig, Yuanyou Li, and Davide Scaramuzza.
\newblock Time lens: Event-based video frame interpolation.
\newblock In {\em IEEE Conf. Comput. Vis. Pattern Recog.}, pages 16155--16164, 2021.

\end{thebibliography}
}

\end{document}